\documentclass[journal]{IEEEtran}
\usepackage{amsmath,amsfonts}
\usepackage{algorithmic}
\usepackage{algorithm}
\usepackage{array}
\usepackage[caption=false,font=normalsize,labelfont=sf,textfont=sf]{subfig}
\usepackage{textcomp}
\usepackage{stfloats}
\usepackage{url}
\usepackage{verbatim}
\usepackage{graphicx}
\usepackage{cite}
\newtheorem{theorem}{Theorem}
\newtheorem{definition}{Definition}
\newtheorem{proposition}{Proposition}
\newtheorem{assumption}{Assumption}
\newtheorem{conjecture}{Conjecture}

\newtheorem{lemma}{lemma}
\usepackage{multirow}
\usepackage{makecell}
\usepackage{amssymb}

\begin{document}

\title{Progressive Feedforward Collapse of \\ ResNet Training}

\author{Sicong Wang, Kuo Gai and Shihua Zhang
% <-this % stops a space
\thanks{This work was supported by the National Key Research and Development Program of China [No. 2019YFA0709501], the CAS Project for Young Scientists in Basic Research [No. YSBR-034], the National Natural Science Foundation of China [Nos. 32341013, 12326614, 12126605]. (The first two authors contributed equally to this work.) (Corresponding author: Shihua Zhang.)}
\thanks{The authors are with the Academy of Mathematics and Systems Science, Chinese Academy of Sciences, Beijing 100190, China, and also with the School of Mathematical Sciences, University of Chinese Academy of Sciences, Beijing 100049, China (E-mail: zsh@amss.ac.cn).}% <-this % stops a space
}

%% The paper headers
%\markboth{IEEE TRANSACTIONS ON NEURAL NETWORKS AND LEARNING SYSTEMS,~Vol.~X, No.~X, X~202X}%
%% Remember, if you use this you must call \IEEEpubidadjcol in the second
%% column for its text to clear the IEEEpubid mark.

\maketitle

\begin{abstract}
Neural collapse (NC) is a simple and symmetric phenomenon for deep neural networks (DNNs) at the terminal phase of training, where the last-layer features collapse to their class means and form a simplex equiangular tight frame aligning with the classifier vectors. However, the relationship of the last-layer features to the data and intermediate layers during training remains unexplored. To this end, we characterize the geometry of intermediate layers of ResNet and propose a novel conjecture, \textit{progressive feedforward collapse} (PFC), claiming the degree of collapse increases during the forward propagation of DNNs. We derive a transparent model for the well-trained ResNet according to that ResNet with weight decay approximates the geodesic curve in Wasserstein space at the terminal phase. The metrics of PFC indeed monotonically decrease across depth on various datasets. We propose a new surrogate model, \textit{multilayer unconstrained feature model} (MUFM), connecting intermediate layers by an optimal transport regularizer. The optimal solution of MUFM is inconsistent with NC but is more concentrated relative to the input data. Overall, this study extends NC to PFC to model the collapse phenomenon of intermediate layers and its dependence on the input data, shedding light on the theoretical understanding of ResNet in classification problems.
\end{abstract}


\section{Introduction}\label{aba:sec1}
\IEEEPARstart{D}{eep} neural networks (DNNs) have been applied to various fields and achieved remarkable success in computer vision \cite{he2016deep}, natural language processing \cite{abdel2014convolutional}, bioinformatics \cite{dong2022deciphering}, and so on. One way to explain the power of DNNs is to discover some specific phenomena in deep learning training and theoretically analyze them. For example, over-parameterized DNNs can achieve zero training error even for random label data. Zhang \textit{et al.} illustrate this phenomenon by a theoretical construction of a two-layer network \cite{zhang2021understanding}. Information bottleneck (IB) theory reveals that the stochastic gradient descent (SGD) optimization has two main phases, i.e., empirical error minimization and representation compression, and the converged layers are closed to the IB bound \cite{tishby2015deep,shwartz2017opening}, which has motivated to apply the IB principle to DNNs training \cite{alemi2016deep,kolchinsky2019nonlinear,9779314}.
Implicit regularization shows that, although there are lots of global minima due to over-parameterization, DNNs trained by SGD still converge to the max-margin solutions without any explicit regularizer \cite{neyshabur2014search,soudry2018implicit,lyu2019gradient}. 

Recently, Papyan \textit{et al.} revealed the neural collapse (NC) phenomenon for a DNN classification task on a balanced dataset, where the last-layer features and classifier vectors exhibit a simple and highly symmetric geometry at the terminal phase of training \cite{papyan2020prevalence}. Specifically, the features collapse to their corresponding class means, while the class means (centered at the global mean) converge to a simplex equiangular tight frame (ETF). Concurrently, the classifier vectors align with the centered class means. As a consequence of this geometry, the classifier converges to choose whichever class has the nearest train class mean. Notably, this phenomenon has been observed across various DNNs employed for classification tasks \cite{han2021neural}.

However, NC solely characterizes the last-layer features and classifier. To gain a more comprehensive understanding of DNNs, we aim to investigate the geometry of intermediate layers. How do DNNs transform the input data to the final simplex ETF during forward propagation? Several studies have indicated that the features become progressively more concentrated and separated across layers, accompanied by a decrease in feature variability through layers \cite{papyan2020traces,zarka2020separation,wang2024understanding,hui2022limitations,galanti2021role,ben2022nearest,MinimalDepth,tirer2023perturbation,he2023law}. Furthermore, examining this problem from a dynamical viewpoint during forward propagation contributes to a clear understanding. Haber and Ruthotto \cite{haber2017stable} and E \cite{weinan2017proposal} first interpreted DNNs with ordinary differential equations. In particular, the residual neural network (ResNet) \cite{he2016deep} stands out as a widely adopted architecture with rich theoretical insights \cite{gai2021mathematical,zhang2021towards,8984747,liu2019towards}. Gai and Zhang \cite{gai2021mathematical} established a connection between the forward propagation of ResNet and the continuity equation, demonstrating that ResNet with weight decay attempts to learn the geodesic curve in the Wasserstein space. In other words, the forward propagation of ResNet is approximately along a straight line at the terminal phase of training, and we call this the geodesic curve assumption.
\IEEEpubidadjcol

\begin{figure*}[!t]
\centering
\includegraphics[width=0.90\textwidth]{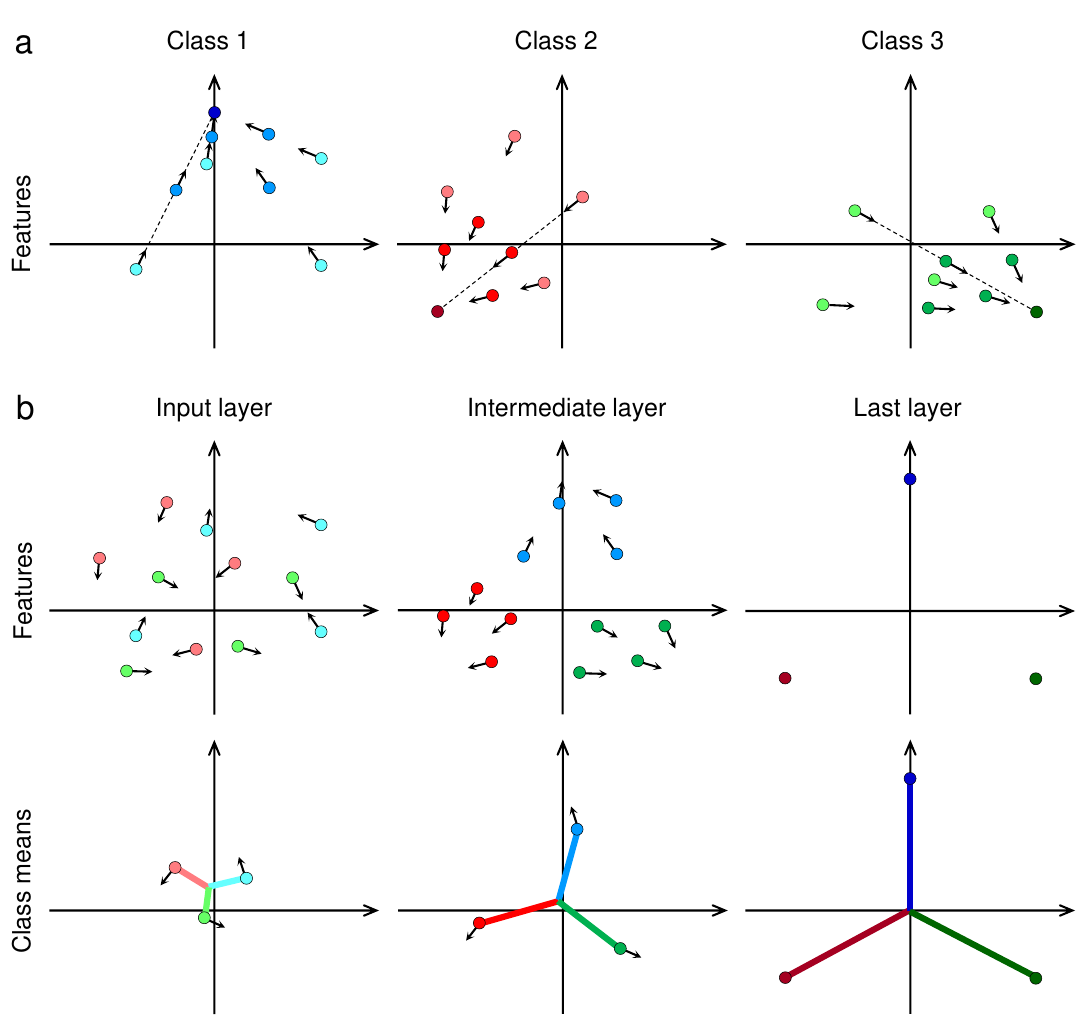}
\caption{\textbf{Illustration of PFC.} This $2$D example represents a three-class dataset, each class depicted by a different color, consisting of four data points per class. Under the geodesic curve assumption, these points undergo uniform linear motion during forward propagation in ResNet, whose color gradually darkens. The arrow indicates the direction of the motion.
\textbf{(a):} Trajectory of features for each class during forward propagation in ResNet. The dotted line shows the motion trajectory. 
\textbf{(b):} \textbf{Left:} Distribution of input. \textbf{Middle:} Distribution of the intermediate layer. \textbf{Right:} Distribution of the last layer, exhibiting NC.
\textbf{Top:} Trajectory of features during forward propagation in ResNet. The features progressively collapse to their respective class means. 
\textbf{Bottom:} Trajectory of (centered) class means. The class means, centered at the global mean, progressively collapse to the simplex ETF (see Definition \ref{ETF}). 
}
\label{figure1}
\end{figure*}

In this paper, we investigate the intermediate layers of ResNet from the perspective of NC and introduce a novel conjecture named progressive feedforward collapse (PFC) (Figure \ref{figure1}). In short, for a well-trained ResNet, the degree of collapse increases during the forward propagation. We empirically show that the PFC conjecture is valid on various datasets, as all metrics associated with PFC decrease through layers at the final epoch. To understand the PFC phenomenon, we incorporate the geodesic curve assumption to model forward propagation and predict the features of intermediate layers. That suggests that ResNets transform input data to the simplex ETF along a straight line, and the features become more collapsed in deeper layers. Theoretically, we can prove that the PFC metrics decrease monotonically while the features collapse to the final simplex ETF along this line under some mild assumptions.

To better illustrate the PFC phenomenon, we propose a novel surrogate model termed the multilayer unconstrained feature model (MUFM) to capture both NC and PFC. Previous studies have theoretically analyzed NC using a simplified model known as unconstrained feature model (UFM) or layer peeled model \cite{mixon2020neural,fang2021exploring,lu2020neural,weinan2022emergence,ji2021unconstrained,han2021neural,tirer2022extended,zhu2021geometric,zhou2022optimization,yaras2022neural,zhou2022all}. This model treats the last-layer features as free variables and optimizes the classifier and features simultaneously, achieving global optimality consistent with NC. However, UFM neglects the relationship between input data and intermediate features, presenting an overidealized scenario. Input data can determine the last-layer features in NC \cite{pmlr-v202-yang23m}. In addition, achieving perfect NC (e.g., all metrics of NC reach zero) is impractical. To connect input data and last-layer features, our proposed MUFM model treats intermediate-layer features as optimization variables through the optimal transport regularizer, serving as a lower bound of weight decay \cite{gai2021mathematical}. The last-layer features are not treated as free variables but are related to data. We demonstrate that the optimal solution of MUFM deviates from the NC solution but exhibits a higher concentration than the initial data. Moreover, there is a trade-off between data and simplex ETF by controlling the coefficient of the optimal transport regularizer.

This paper is organized as follows: Section \ref{sec2} provides preliminaries, including notations, the background of NC, and the theoretical understanding of NC and ResNet. In Section \ref{sec3}, we propose the PFC conjecture and elucidate it under the geodesic curve assumption. Subsequently, we propose MUFM to understand both NC and PFC. In Section \ref{sec4}, we provide some empirical evidence for the conjecture and results mentioned in Section \ref{sec3}. In Section \ref{sec5}, we list some related works. Finally, Section \ref{sec6} provides conclusions and discussions.

\section{Preliminaries}\label{sec2}
In this paper, we focus on DNNs trained on balanced datasets. We consider a $K$-class classification problem, where each class comprises $n$ training samples, i.e. $nK$ samples in total. Denote $\boldsymbol{x}_{k, i} \in \mathbb{R}^D$ as the $i$-th training sample of the $k$-th class. Its corresponding label is the one-hot vector $\boldsymbol{y}_k \in \mathbb{R}^K$.
Formally, DNN can be written as
\begin{align*}
f_{\boldsymbol{\Theta}}(\boldsymbol{x})&=\boldsymbol{W h}_{\boldsymbol{\theta}}(\boldsymbol{x})+\boldsymbol{b}, \\
\boldsymbol{h}_{\boldsymbol{\theta}}(\boldsymbol{x})&=\sigma\left(\boldsymbol{W}_{L}\left(\cdots \sigma\left(\boldsymbol{W}_1 \boldsymbol{x}+\boldsymbol{b}_1\right) \right)+\boldsymbol{b}_{L}\right), 
\end{align*}
where $\boldsymbol{h}_{\boldsymbol{\theta}}(\cdot): \mathbb{R}^D \rightarrow \mathbb{R}^d$ is the feature mapping, $\sigma(\cdot)$ is an element-wise nonlinear function. $\boldsymbol{W}=\left[\boldsymbol{w}_1, \cdots, \boldsymbol{w}_K\right]^{\top} \in \mathbb{R}^{K \times d}$ and $\boldsymbol{b} \in \mathbb{R}^K$ are the classifier matrix and bias. Specifically, we assume $d>K$ which is a common scenario in practice. $\boldsymbol{\Theta}=\{\boldsymbol{W}, \boldsymbol{b}, \boldsymbol{\theta}\}$ is the set of all DNN parameters.

Our goal is to train the parameters $\boldsymbol{\Theta}$ by minimizing the empirical risk over all training samples,
$$
\min _{\boldsymbol{\Theta}} \frac{1}{K n} \sum_{k=1}^K \sum_{i=1}^n \mathcal{L}\left(f_{\boldsymbol{\Theta}}(\boldsymbol{x}_{k,i}), \boldsymbol{y}_k\right)+\mathcal{R}(\boldsymbol{\Theta}),
$$
where $\mathcal{L}(\cdot, \cdot)$ is the loss function, e.g., cross-entropy (CE) or mean squared error (MSE), and $\mathcal{R}(\cdot)$ is the regularizer (e.g., $L_2$-norm regularization). 

For $l=1,\cdots, L$, let's denote the $l$-layer feature of the $i$-th training sample of the $k$-th class by $\boldsymbol{h}^l_{k, i}$ and the last-layer features by $\{\boldsymbol{h}_{k, i}\}$ (i.e., $\boldsymbol{h}_{k, i} =\boldsymbol{h}^L_{k, i}= \boldsymbol{h}_{\boldsymbol{\theta}}(\boldsymbol{x}_{k, i})$). The class mean and global mean of the $l$-layer features are $\boldsymbol{h}^l_{k}$ and $\boldsymbol{h}^l_{G}$,
\begin{align*}
\boldsymbol{h}^l_{k} &= \frac{1}{n}\sum_{i=1}^{n}\boldsymbol{h}^l_{k,i}, \\ 
\boldsymbol{h}^l_{G} &= \frac{1}{K}\sum_{k=1}^{K}\boldsymbol{h}^l_{k}= \frac{1}{nK}\sum_{k=1}^{K}\sum_{i=1}^{n}\boldsymbol{h}^l_{k,i}.
\end{align*}
The $l$-layer features within-class covariance $\boldsymbol{\Sigma}_W^l$ and between-class covariance $\boldsymbol{\Sigma}_B^l$ are computed by
\begin{align*}
\boldsymbol{\Sigma}_W^l &= \frac{1}{nK}\sum_{k=1}^{K}\sum_{i=1}^{n}\left(\boldsymbol{h}_{k, i}^l-\boldsymbol{h}_k^l\right)\left(\boldsymbol{h}_{k, i}^l-\boldsymbol{h}_k^l\right)^{\top}, \\ 
\boldsymbol{\Sigma}_B^l &=\frac{1}{K}\sum_{k=1}^{K}\left(\boldsymbol{h}_k^l-\boldsymbol{h}_G^l\right)\left(\boldsymbol{h}_k^l-\boldsymbol{h}_G^l\right)^{\top}.
\end{align*}

\subsection{Neural Collapse}
Papyan \textit{et al.} revealed the NC phenomenon for a DNN classification task on a balanced dataset, where the last-layer features and classifier vectors exhibit a simple and highly symmetric geometry at the terminal phase of training. We introduce simplex ETF first. 
\begin{definition}[Simplex ETF]\label{ETF}
A simplex ETF is a collection of vectors $\boldsymbol{m}_i \in \mathbb{R}^d, i=$ $1,2, \cdots, K, d \geq K-1$ specified by the columns of
\begin{equation}
\boldsymbol{M}=\sqrt{\frac{K}{K-1}} \boldsymbol{U}\left(\boldsymbol{I}_K-\frac{1}{K} \boldsymbol{1}_K \boldsymbol{1}_K^T\right),
\end{equation}
where $\boldsymbol{U} \in \mathbb{R}^{d \times K}$ is a partial orthogonal matrix and satisfies $\boldsymbol{U}^T \boldsymbol{U}=\boldsymbol{I}_K, \boldsymbol{I}_K\in \mathbb{R}^{K \times K}$ is the identity matrix, and $\boldsymbol{1}_K\in \mathbb{R}^K$ is the all ones vector. 
\end{definition}
Formally, NC can be described with four properties:

(NC1) \textbf{Variability collapse}: the within-class variability converges to zero, which means the last-layer features of each class collapse to their class means during the training, i.e.,
$$\boldsymbol{\Sigma}_W \to \boldsymbol{0}.$$

(NC2) \textbf{Convergence to the simplex ETF}: the class means centered at global mean converge to the simplex ETF (see Definition \ref{ETF}). That is to say, the centered class means have equal length and equal, maximally separated pair-wise angles, i.e.,
$$
\begin{gathered}
\frac{\left\langle\boldsymbol{h}_k-\boldsymbol{h}_G, \boldsymbol{h}_{k^{\prime}}-\boldsymbol{h}_G\right\rangle}{\left\|\boldsymbol{h}_k-\boldsymbol{h}_G\right\|_2\left\|\boldsymbol{h}_{k^{\prime}}-\boldsymbol{h}_G\right\|_2} \rightarrow \begin{cases}1, & k=k^{\prime} \\
\frac{-1}{K-1}, & k \neq k^{\prime}\end{cases} \\
\left\|\boldsymbol{h}_k-\boldsymbol{h}_G\right\|_2-\left\|\boldsymbol{h}_{k^{\prime}}-\boldsymbol{h}_G\right\|_2 \rightarrow 0, \quad \forall k \neq k^{\prime}.
\end{gathered}
$$

(NC3) \textbf{Convergence to self-duality}: the classifier vectors are in alignment with the centered class means, i.e.,
$$
\dfrac{\boldsymbol{w}_k}{\left\|\boldsymbol{w}_k\right\|_2} - \dfrac{\boldsymbol{h}_k-\boldsymbol{h}_G}{\left\|\boldsymbol{h}_k-\boldsymbol{h}_G\right\|_2} \rightarrow 0, \quad \forall k.
$$

(NC4) \textbf{Simplification to nearest class center (NCC)}: the classifier converges to choosing whichever class has the nearest train class mean. Given a new feature $\boldsymbol{h}$, it is classified by 
$$
\arg \max _{k}\left\langle\boldsymbol{w}_{k}, \boldsymbol{h}\right\rangle+b_{k} \rightarrow \underset{k}{\arg \min }\left\|\boldsymbol{h}-\boldsymbol{h}_{k}\right\|_2.
$$

The NC phenomenon only characterizes the geometry of the last-layer features. It is natural to investigate the behavior of intermediate layers. Several works have shown that the features become more and more concentrated and separated through layers \cite{papyan2020traces,zarka2020separation,wang2024understanding}. Furthermore, many empirical findings indicate that the variability collapse (NC1) emerges at intermediate layers but weaker than the last layer \cite{hui2022limitations,galanti2021role,ben2022nearest,MinimalDepth,tirer2023perturbation,he2023law}. Ben-Shaul and Dekel \cite{ben2022nearest}, Galanti \textit{et al.} \cite{MinimalDepth} also provided a similar result of NC4. To characterize NC at the intermediate layers more carefully, we introduce the terminology \textit{nearest class-center (NCC) separable} and \textit{empirical effective depth} proposed in \cite{MinimalDepth}.

That the $l$-layer features are NCC separable means the true class of each sample can be identified correctly by finding the nearest class mean of the corresponding feature. Formally, for $\forall k\in [K], i \in [n]$, we have
$$\underset{c \in[K]}{\arg \min }\left\|\boldsymbol{h}_{k,i}^l-\boldsymbol{h}_{c}^l\right\|_2=k.$$
Next, we use the empirical effective depth to denote the lowest depth that satisfies NCC separability in practice. The empirical result shows that if the $L_0$-layer features are NCC separable, then all $l$-layer ($l\ge L_0$) features are NCC separable. We allow a small error $\epsilon$ when measuring the NCC separability empirically. Consider a $L$-hidden layer well-trained neural network, the $\epsilon$-empirical effective depth is the minimal $l\in [L]$ that satisfies, 
$$\frac{1}{nK}\sum_{k=1}^{K}\sum_{i=1}^{n} \mathbb{I}\left[\underset{c \in[K]}{\arg \min }\left\|\boldsymbol{h}_{k,i}^l-\boldsymbol{h}_{c}^l\right\|_2 \neq k\right]\le \epsilon.$$

Recently, several works characterized the intermediate NC properties after effective depth, e.g., the layers after a certain layer exhibit NC properties \cite{rangamani2023feature,sukenik2024deep}. We mainly focus on the layers before effective depth in this paper. As an intriguing phenomenon, several works also investigated some implications of NC for DNNs performance \cite{yang2022inducing,xie2023neural,pmlr-v206-liu23i,haas2023linking} (see a review for more details \cite{kothapalli2022neural}).

\subsection{Understanding NC by an Unconstrained Feature Model}
It is challenging to characterize the NC phenomenon in theory due to the highly non-convex of DNNs with the interaction between different layers. Thus, previous works only explored a simplified model (i.e., an unconstrained feature model (UFM) or layer-peeled model) instead of DNNs. 

UFM treats the last-layer features as free optimization variables and neglects the intermediate layers, so the DNN is reduced to a linear model whose variables are the last-layer features and the classifier vectors. This simplification is based on the fact that DNNs are highly overparameterized and can approximate a large family of functions. We can focus on the last-layer features $\boldsymbol{H}$ and the classifier $\boldsymbol{W}$ to study their optimality. Researchers mainly study the global optimality, loss landscape, and training dynamics in different settings of UFM, such as different loss functions (e.g., MSE, CE), regularizers (weight decay), or constraints on $\boldsymbol{W}$ and $\boldsymbol{H}$. For example, we introduce the UFM under the cross-entropy loss and weight decay on $\boldsymbol{W}$ and $\boldsymbol{H}$.

Let $\boldsymbol{W}=\left[\boldsymbol{w}_1, \boldsymbol{w}_2, \cdots, \boldsymbol{w}_K\right]^{\top} \in \mathbb{R}^{K \times d}$ and $\boldsymbol{H}=\left[\boldsymbol{h}_{1,1}, \cdots, \boldsymbol{h}_{1, n}, \boldsymbol{h}_{2,1}, \cdots, \boldsymbol{h}_{K, n}\right] \in \mathbb{R}^{d \times K n}$ be the matrices of classifier and last-layer features, UFM is defined as follows:
\begin{equation}
\min _{\boldsymbol{W}, \boldsymbol{H}} \frac{1}{K n} \sum_{k=1}^{K} \sum_{i=1}^{n} \mathrm{CE}\left(\boldsymbol{W} \boldsymbol{h}_{k, i}, \boldsymbol{y}_{k}\right)+\frac{\lambda_{\boldsymbol{W}}}{2}\|\boldsymbol{W}\|_{F}^{2}+\frac{\lambda_{\boldsymbol{H}}}{2}\|\boldsymbol{H}\|_{F}^{2},
\end{equation}
where $\mathrm{CE}$ indicates the cross-entropy loss, and $\lambda_{\boldsymbol{W}}, \lambda_{\boldsymbol{H}}>0$ are the hyperparameters for weight decay.

Although the problem is non-convex, recent works proved that the global minimizers are indeed NC solutions \cite{lu2020neural,weinan2022emergence,fang2021exploring,ji2021unconstrained,tirer2022extended,zhu2021geometric,zhou2022optimization,yaras2022neural,zhou2022all}. Moreover, some studies \cite{zhu2021geometric,zhou2022optimization,yaras2022neural,zhou2022all} showed that UFM has benign loss landscape, that is there is no spurious local minimizer on the landscape, which means all local minima are global minima. Therefore, any optimization methods that can escape strict saddle points (e.g., stochastic gradient descent) will converge to NC solutions. Mixon \textit{et al.} \cite{mixon2020neural}, Han \textit{et al.} \cite{han2021neural} studied the gradient flow of UFM, and proved that it converges to NC solutions under some assumptions. Beyond UFM, some studies extended it to two or arbitrary layers and proved the global optimality \cite{tirer2022extended,sukenik2024deep,dang2023neural}, which is closer to the practical DNNs.

Overall, as a simplified model of DNNs, UFM is consistent with NC. It gives a theoretical understanding of it but is still limited. This simplification breaks the relation between the data and the learned features, and the regularizer or constraint on $\boldsymbol{H}$ is not used in practice. The regularizer on $\boldsymbol{H}$ is interpreted as upper bounded by the weight decay on all layers if the inputs are bounded. 

\subsection{ResNet Learns the Geodesic Curve in the Wasserstein Space}\label{aba:sec2.3}
NC only characterizes the last-layer features and UFM breaks the relation between data and features. To explore more layers and take forward propagation into account, we introduce the dynamical system view of deep learning in this subsection. Gai and Zhang \cite{gai2021mathematical} built the connection of ResNet and the continuity equation, showing that ResNet with weight decay attempts to learn the geodesic curve in the Wasserstein space. Thus, the forward propagation of ResNet is along a straight line.

The connection between ResNet and the dynamical system was first introduced in \cite{haber2017stable,weinan2017proposal}. The general form of a ResNet block \cite{he2016identity} in forward propagation is
\begin{equation}
\boldsymbol{x}^{(l+1)}=\boldsymbol{x}^{(l)}+v_l\left(\boldsymbol{x}^{(l)}\right), l \in\{0,1, \cdots, L-1\}
\end{equation}
where $v_l(\cdot)$ is a function induced by a shallow network, and $\boldsymbol{x}^{(l)}$ is the input of the $l$-th ResNet block. It can be viewed as a discretization of the dynamical system
\begin{equation}
\frac{d \boldsymbol{x}}{d t}=v(\boldsymbol{x}, t), t \in[0,1]
\end{equation}

Denote the distribution of the $k$-th ResNet block output, data distribution and target distribution by $\mu_k,\mu_0,\mu_1$, respectively. Then the corresponding al system has the form
\begin{equation}\label{dynamic system}
\frac{d \mu_t}{d t}=\tilde{v}\left(\mu_t, t\right), t \in[0,1]
\end{equation}
Gai and Zhang \cite{gai2021mathematical} claimed that the dynamical system (\ref{dynamic system}) should conserve mass, corresponding to the continuity equation which has the form
\begin{equation}\label{continuity equation}
\frac{d \mu_t}{d t}+\nabla \cdot\left(v_t \mu_t\right)=0, t \in[0,1]
\end{equation}
where the vector field $v_t$ is the infinitesimal variation of the continuum $\mu_t$.

When $\mu_0$ and $\mu_1$ are fixed, there are infinite curves in the probability measure space connecting $\mu_0$ and $\mu_1$ by continuity equations. It has been shown that a special one, the geodesic curve in the Wasserstein space has good generalization ability. It is induced by the optimal transport map $T$ from $\mu_0$ to $\mu_1$ with the minimum cost. This cost is called the Wasserstein distance $W_c\left(\mu_0, \mu_1\right)$, i.e.
$$
W_c\left(\mu_0, \mu_1\right)=\inf _{T_{\#} \mu_0=\mu_1} \int c(\boldsymbol{x}, T(\boldsymbol{x})) d \mu_0(x),
$$
where $c(\cdot, \cdot)$ is the cost function of transporting mass from $\mu_0$ to $\mu_1$, $T_{\#}$ is the push-forward operator. The cost function is usually set to $\|\boldsymbol{x}-T(\boldsymbol{x})\|_2^2$ in both optimal transport and deep learning. In this setting, the geodesic curve $\left(\mu_t\right)$ is induced by the optimal transport map $T$, i.e.
$$
\mu_t=\left((1-t) I_d+t T\right)_{\#} \mu_0 .
$$

For each point $\boldsymbol{x} \in \operatorname{supp}\left(\mu_0\right), \boldsymbol{x}$ is transported along a straight line $\boldsymbol{x}^{(t)}=(1-t) \boldsymbol{x}+t T(\boldsymbol{x})$. 
Thus, we want to get the optimal transport map $T$ by the neural network, then the forward propagation is along a straight line. The following proposition provides a way to compute the Wasserstein distance.

\begin{proposition}(Benamou-Brenier formula\cite{benamou1999numerical})
Let $\mu_0$, $\mu_1$ $\in$ $\mathcal{P}\left(\mathbb{R}^d\right)$. Then it holds
$$
W_2^2\left(\mu_0, \mu_1\right)=\inf \left\{\int_0^1\left\|v_t\right\|^2_{L^2\left(\mu_t\right)} d t\right\},
$$
where the infimum is taken among all weakly continuous distributional solutions of the continuity equation $\left(\mu_t, v_t\right)$ connecting $\mu_0$ and $\mu_1$.
\end{proposition}

Gai and Zhang \cite{gai2021mathematical} proved that the weight decay is an upper bound of the above energy in DNNs. In addition, the forward propagation of ResNet satisfies the continuity equation. According to the Benamou-Brenier formula, ResNet with weight decay aims to approximate the geodesic curve in the Wasserstein space. Thus, we propose the geodesic curve assumption,
\begin{assumption}(Geodesic curve assumption)
At the terminal phase of training, the ResNet with weight decay has learned the geodesic curve in the Wasserstein space $\mathcal{P}\left(\mathbb{R}^d\right)$, which is induced by the optimal transport map. As a result, the forward propagation of ResNet is along a straight line in the feature space $\mathbb{R}^d$.
\end{assumption}

\section{Progressive Feedforward Collapse}\label{sec3}
NC reveals a distinctive geometric structure in the last-layer features of DNNs. It is interesting to investigate the behavior of features in intermediate layers, particularly how neural networks like ResNet transform the initial data to the simplex ETF during forward propagation. This process is crucial for gaining insights into its classification tasks. While previous studies have shown that features tend to become increasingly concentrated and separated through layers \cite{papyan2020traces,zarka2020separation,wang2024understanding,hui2022limitations,galanti2021role,ben2022nearest,MinimalDepth,tirer2023perturbation,he2023law}, a clear and transparent model of the entire network, analogous to the NC phenomenon observed in the last layer, is still lacking. To address this issue, in this section, we mainly consider the well-trained ResNet and propose a novel conjecture named progressive feedforward collapse (PFC).

\subsection{PFC Conjecture}\label{sec3.1}
To investigate the behavior of features in intermediate layers, we first extend the statistics used in NC to intermediate layers. For features of each layer, we compute the following metrics to measure the degree of collapse.

The variability collapse of the layer $l$:
\begin{equation}
\mathcal{PFC} _1(\boldsymbol{H}^{l}) := \frac{\mathbb{E}\left[\left\|\boldsymbol{h}_{k,i}^l-\boldsymbol{h}_{k}^l\right\|_2^2\right]}{\mathbb{E}\left[\left\|\boldsymbol{h}_{k}^l-\boldsymbol{h}_G^l\right\|_2^2\right]} = 
\dfrac{\mathrm{Tr}(\boldsymbol{\Sigma}_W^l)}{\mathrm{Tr}(\boldsymbol{\Sigma}_B^l)}.
\end{equation}
We use the same metric in \cite{hui2022limitations,tirer2023perturbation}, which is a little different from the original NC1 metric $\text{Tr}(\boldsymbol{\Sigma}_W\boldsymbol{\Sigma}_B^{\dagger})$ \cite{papyan2020prevalence}, 
where $\boldsymbol{\Sigma}_B^{\dagger}$ denotes the pseudo-inverse of $\boldsymbol{\Sigma}_B$. The numerator of $\mathcal{PFC} _1^{(l)}$ is the within-class variance and the denominator is the between-class variance. $\mathcal{PFC} _1^{(l)}$ is more amenable for analysis. It can still capture the behavior of variability collapse that is $\boldsymbol{\Sigma}_W \to \boldsymbol{0}$ while $\boldsymbol{\Sigma}_B \not\to \boldsymbol{0}$ is non-degenerate. 

The distance between the $l$-layer (centered) class means and the simplex ETF:
\begin{equation}
\mathcal{PFC} _2(\boldsymbol{H}^{l}):=\left\|\frac{\tilde{\boldsymbol{H}^{l}}^{\top} \tilde{\boldsymbol{H}^{l}}}{\left\|\tilde{\boldsymbol{H}^{l}}^{\top} \tilde{\boldsymbol{H}^{l}}\right\|_F}-\boldsymbol{E}\right\|_F,
\end{equation}
where $
\tilde{\boldsymbol{H}^{l}}:=\left[
\boldsymbol{h}_1^{l}-\boldsymbol{h}_G^{l}, \cdots, \boldsymbol{h}_K^{l}-\boldsymbol{h}_G^{l}
\right] \in \mathbb{R}^{d \times K}
$ is the centered class mean matrix, $\boldsymbol{E}=\frac{1}{\sqrt{K-1}}\left(\boldsymbol{I}_K-\frac{1}{K} \boldsymbol{1}_K \boldsymbol{1}_K^{\top}\right)$. We extend the metric in \cite{zhu2021geometric} to each layer, considering the geometry of simplex ETF with a single metric.

The nearest class center (NCC) accuracy of the layer $l$:
\begin{equation}
\mathcal{PFC} _3(\boldsymbol{H}^{l}):=\frac{1}{nK}\sum_{k=1}^{K}\sum_{i=1}^{n} \mathbb{I}\left[\underset{c \in[K]}{\arg \min }\left\|\boldsymbol{h}_{k,i}^l-\boldsymbol{h}_{c}^l\right\|_2 = k\right].
\end{equation}
$\mathcal{PFC}_3^{(l)}$ measures the NCC separability of each layer. When $\mathcal{PFC}_3^{(l)} = 1$, we say that the features of the layer $l$ are NCC separable. Intuitively, the features are more NCC separable when they are closer to their class mean and the centered class means are closer to the simplex ETF.

At the terminal phase of training, the last-layer features collapse to their class means and the class means (centered at global mean) converge to the simplex equiangular tight frame (ETF). Meanwhile, DNNs transform the initial data to the simplex ETF during forward propagation. The metrics mentioned above can be employed to quantify the degree of collapse. By computing these metrics at each layer, we can investigate the geometry of intermediate layers and observe the variation trend. As a result, we can conjecture that 
\begin{conjecture}[Progressive Feedforward Collapse conjecture]
At the terminal phase of ResNet training, NC emerges at the last-layer features. There exists an order in the degree of collapse (measured by three PFC metrics) of each layer before effective depth. Concretely,

\textbf{(PFC1)} The features of each layer progressively collapse to their class means.

\textbf{(PFC2)} The centered class means of each layer features progressively collapse to the simplex ETF.

\textbf{(PFC3)} The NCC accuracy of each layer progressively collapses to $1$.

During the training process, the metrics at each layer start at a random initialization and gradually converge to the final order.
\end{conjecture}

The PFC conjecture characterizes the variation of geometry in the forward propagation of ResNet. It shows that each ResNet block makes the features more concentrated to their class means and the class means more separated. Thus, ResNet progressively extracts features during forward propagation that make data classified more easily.
We give partial proof of this conjecture under the geodesic curve assumption in the next subsection and some empirical evidence in Section \ref{sec4}.

\subsection{Monotonicity under the Geodesic Curve Assumption}
In Section \ref{aba:sec2.3}, we introduce the geodesic curve assumption, that is the forward propagation of ResNet is along a straight line. Formally, for each data point $\boldsymbol{x}_{k, i}$, the feature map in the forward propagation of ResNet $\boldsymbol{h}_{k, i}^{(t)}, t \in [0,1]$ is along the line $\boldsymbol{h}_{k, i}^{(t)}=(1-t) \boldsymbol{h}_{k, i}^{(0)}+t \boldsymbol{h}_{k, i}^{(1)}, t \in [0,1]$. 
This assumption models the forward propagation and can help to reveal how ResNet transforms the initial data to the simplex ETF. PFC shows that the features progressively collapse to the final simplex ETF. In this section, we prove that under the geodesic curve assumption, PFC metrics monotonically decrease across depth at the terminal phase of training which explains the PFC phenomenon. The detailed proofs are provided in Appendix \ref{appxA} and \ref{appxB}.

Suppose the last-layer features $\{\boldsymbol{h}_{k, i}^{(1)}\}$ exhibit NC, that is $\boldsymbol{h}_{k, i}^{(1)} = \boldsymbol{h}_{k}^{(1)}, \forall i,k,$ and the set of vectors $\{\boldsymbol{h}_{k}^{(1)} - \boldsymbol{h}_{G}^{(1)}\}$ composes the simplex ETF. Then, under the geodesic curve assumption, the forward propagation of ResNet is 
\begin{equation}\label{line}
\begin{aligned}
\boldsymbol{h}_{k, i}^{(t)} &=(1-t) \boldsymbol{h}_{k, i}^{(0)}+t \boldsymbol{h}_{k}^{(1)}, t \in [0,1] \\
\boldsymbol{h}_{k}^{(t)} &= (1-t) \boldsymbol{h}_{k}^{(0)}+t \boldsymbol{h}_{k}^{(1)}, t \in [0,1] \\
\boldsymbol{h}_{G}^{(t)} &= (1-t) \boldsymbol{h}_{G}^{(0)}+t \boldsymbol{h}_{G}^{(1)}, t \in [0,1] 
\end{aligned}
\end{equation}
the forward propagation of centered class mean matrix of ResNet $\tilde{\boldsymbol{H}}(t):=\left[
\boldsymbol{h}_1^{(t)}-\boldsymbol{h}_G^{(t)},\cdots,\boldsymbol{h}_K^{(t)}-\boldsymbol{h}_G^{(t)}
\right]$ is 
\begin{equation}\label{line2}
\tilde{\boldsymbol{H}}(t) = (1-t)\tilde{\boldsymbol{H}}(0)+t\tilde{\boldsymbol{H}}(1), t \in [0,1]
\end{equation}

\begin{theorem}\label{thm1}
Given the start and end points $\{\boldsymbol{h}_{k, i}^{(0)}\}$ and $\{\boldsymbol{h}_{k, i}^{(1)}\}$ as above, assume that $\sum_{k=1}^{K}\langle \boldsymbol{h}_{k}^{(0)} - \boldsymbol{h}_{G}^{(0)},\boldsymbol{h}_{k}^{(1)} - \boldsymbol{h}_{G}^{(1)}\rangle \ge 0$, along the line (\ref{line}), PFC1 metric monotonically decreases to zero in $[0,1]$.
\end{theorem}

Given the input $\{\boldsymbol{h}_{k}^{(0)}\}$, ResNet transforms $\{\boldsymbol{h}_{k, i}^{(0)}\}$ to $\{\boldsymbol{h}_{k}^{(1)}\}$ as a optimal transport map with the minimum cost. That is, the transport cost $\sum_{k=1}^{K}\|\tilde{\boldsymbol{h}}_{k}^{(0)} - \tilde{\boldsymbol{h}}_{k}^{(1)}\|_2^2 $ is relatively small. Thus, it is reasonable for the input data to satisfy the assumption
\begin{equation*}
\begin{aligned}
&\sum_{k=1}^{K}\langle \boldsymbol{h}_{k}^{(0)} - \boldsymbol{h}_{G}^{(0)},\boldsymbol{h}_{k}^{(1)} - \boldsymbol{h}_{G}^{(1)}\rangle \\
=& \sum_{k=1}^{K}\frac{1}{2}\left( \left\|\tilde{\boldsymbol{h}}_{k}^{(0)}\right\|_2^2+\left\|\tilde{\boldsymbol{h}}_{k}^{(1)}\right\|_2^2 - \left\|\tilde{\boldsymbol{h}}_{k}^{(0)} - \tilde{\boldsymbol{h}}_{k}^{(1)}\right\|_2^2\right) \\
=& \frac{1}{2}\sum_{k=1}^{K}\left( \left\|\tilde{\boldsymbol{h}}_{k}^{(0)}\right\|_2^2+\left\|\tilde{\boldsymbol{h}}_{k}^{(1)}\right\|_2^2\right) - \frac{1}{2}\sum_{k=1}^{K}\left\|\tilde{\boldsymbol{h}}_{k}^{(0)} - \tilde{\boldsymbol{h}}_{k}^{(1)}\right\|_2^2 \ge 0.
\end{aligned}
\end{equation*}
Empirical result in Section \ref{sec4.1} shows that this assumption is valid and the PFC1 metric is indeed a monotonic decrease.
\begin{theorem}\label{thm2}
Given the start and end points $\tilde{\boldsymbol{H}}(0)$ and $\tilde{\boldsymbol{H}}(1)$ as above, assume that the transport cost $\|\tilde{\boldsymbol{H}}(1) - \tilde{\boldsymbol{H}}(0)\|_F^2 $ is sufficiently small, $\|\tilde{\boldsymbol{H}}(t)^{\top} \tilde{\boldsymbol{H}}(t)\|_F^2 >0$ along the line (\ref{line2}), we have the PFC2 metric monotonically decreases to zero in $[0,1]$.
\end{theorem}

Intuitively, the features are NCC separable when they are concentrated and separated. That is to say, PFC1 and PFC2 can imply PFC3 implicitly. Thus, the PFC metrics decrease monotonically, corresponding to PFC. Besides geodesic curve assumption, we need the transport cost $\sum_{k=1}^{K}\|\tilde{\boldsymbol{h}}_{k}^{(0)} - \tilde{\boldsymbol{h}}_{k}^{(1)}\|_2^2 $$= \|\tilde{\boldsymbol{H}}(1) - \tilde{\boldsymbol{H}}(0)\|_F^2$ is sufficiently small to induce monotonic decrease. Gai and Zhang \cite{gai2021mathematical} observed that ResNet with weight decay attempts to learn an optimal transport map, and thus relatively small transport cost is reasonable.

\subsection{Multilayer Unconstrained Feature Model}\label{aba:sec3.3}
Analyzing the NC phenomenon directly in DNNs is challenging due to the intricate interactions across multiple layers. Consequently, various studies have opted for simplification using unconstrained feature or layer-peeled models, focusing solely on the last-layer features and the classifier. The unconstrained feature model regards the last-layer features as free variables and optimizes features and the classifier under loss function and regularizer/constraint. The global optimality achieved by UFM aligns with the NC phenomenon.

While UFM is a reasonable surrogate model for understanding NC, it has inherent limitations. First, UFM breaks the relation between the initial data and the features, even though the initial data influences the collapsed features \cite{pmlr-v202-yang23m}. It neglects the structure of DNNs, especially the depth, which holds practical significance. Next, for symmetry, UFM usually uses the regularizer/constraint on $\|\boldsymbol{H}\|_{F}$ and $\|\boldsymbol{W}\|_{F}$ simultaneously. In practice, weight decay is commonly applied to all parameters without explicitly constraining the features. The formulation of UFM is quite idealized. In addition, we usually can not reach perfect NC in practice. To overcome the limitations of UFM, we propose our multilayer unconstrained feature model (MUFM) based on the geodesic curve assumption. 

In addition to the last-layer features, we treat all features $\{\boldsymbol{H}^l\}$$= \{[\boldsymbol{h}_{1, 1}^l,\cdots,\boldsymbol{h}_{1, n}^l, \boldsymbol{h}_{2,1}^l, \cdots,\boldsymbol{h}_{K, n}^l]\}$, $l = 1,\cdots,L$ as optimization variables in MUFM, and connect them by the optimal transport regularizer, which is a lower bound of the weight decay as mentioned in Section \ref{aba:sec2.3}. Formally, recall that the general form of a ResNet block in forward propagation is
\begin{equation*}
\boldsymbol{x}^{(l+1)}=\boldsymbol{x}^{(l)}+v_l\left(\boldsymbol{x}^{(l)}\right), l \in\{0,1, \cdots, L-1\}
\end{equation*}
where $v_l(\cdot)=\sigma\left(\boldsymbol{W}_{l} \cdot\right)$ is a shallow network, $\sigma(\cdot)$ is the ReLU function and $\boldsymbol{x}^{(l)}$ is the input of the $l$-th ResNet block. Let $\mu_l$ denote the distribution of $\boldsymbol{x}^{(l)}$ and assume $\|\boldsymbol{x}^{(l)}\|_2$ is upper bounded. Then, there exists a constant $C$ such that
\begin{align*}
C \sum_{l=0}^{L-1}\left(\left\|\boldsymbol{W}_{l}\right\|_F^2 \right) &\geq \sum\limits_{l=0}^{L-1}\left\|v_l\right\|^2_{L^2(\mu_l)} \\
& = \frac{1}{K n}\sum_{l=0}^{L-1} \sum_{k=1}^K \sum_{i=1}^n \left\|\boldsymbol{h}_{k, i}^{l+1}-\boldsymbol{h}_{k, i}^l\right\|_2^2. 
\end{align*}
In particular, if the input distribution $\mu_l$ is symmetric and isotropic with zero-mean (e.g., standard multivariate normal distribution), then the optimal transport regularizer is equivalent to the weight decay up to a constant factor. Actually, 
\begin{align*}
\left\|v_l\right\|^2_{L^2(\mu_l)} &= \mathbb{E}_{\boldsymbol{x}^{(l)}}\left[\left\|\sigma\left(\boldsymbol{W}_l\boldsymbol{x}^{(l)}\right)\right\|_2^2\right] \\
&= \frac{1}{2} \mathbb{E}_{\boldsymbol{x}^{(l)}}\left[\left\|\boldsymbol{W}_l\boldsymbol{x}^{(l)}\right\|_2^2\right] \\
&= \frac{1}{2}\left\|\boldsymbol{W}_l\right\|_F^2.
\end{align*}
Thus, it is reasonable to replace the weight decay on former layers with the optimal transport regularizer and only consider weight decay on classifier $\boldsymbol{W}$. Consider the following optimization problem, 
\begin{equation}\label{prob}
\begin{aligned}
&
\begin{aligned}
\min _{\boldsymbol{W}, \boldsymbol{H}^1,\dots,\boldsymbol{H}^L} &\mathcal{L}_{CE/MSE}(\boldsymbol{W},\boldsymbol{H}^L,\boldsymbol{Y})+\frac{\lambda_{\boldsymbol{W}}}{2K}\|\boldsymbol{W}\|_F^2 \\
& + \frac{\lambda}{2Kn}\sum_{l=0}^{L-1} \sum_{k=1}^K \sum_{i=1}^n \left\|\boldsymbol{h}_{k, i}^{l+1}-\boldsymbol{h}_{k, i}^l\right\|_2^2 
\end{aligned}\\
&\text { s.t. } \quad \boldsymbol{h}_{k, i}^0=\boldsymbol{x}_{k, i}, \quad \forall k=1,\cdots,K, i = 1,\cdots,n,\\ 
\end{aligned}
\end{equation}
where 
$$
\mathcal{L}_{CE}(\boldsymbol{W},\boldsymbol{H}^L,\boldsymbol{Y}) = \frac{1}{K n}\sum_{k=1}^K \sum_{i=1}^n \mathrm{CE}\left(\boldsymbol{W} \boldsymbol{h}_{k, i}^L, \boldsymbol{y}_k\right),
$$
$$
\mathcal{L}_{MSE}(\boldsymbol{W},\boldsymbol{H}^L,\boldsymbol{Y}) = \frac{1}{2K n} \|\boldsymbol{WH}^L-\boldsymbol{Y}\|_{F}^{2}
$$
and $\otimes$ denotes the Kronecker product, $\lambda_{\boldsymbol{W}}, \lambda>0$ are the hyperparameters of regularizers, $\mathbf{Y}=\mathbf{I}_K \otimes \mathbf{1}_n^{\top} \in \mathbb{R}^{K \times K n}$ for MSE loss.

When minimizing the optimal transport regularizer, if the last-layer features are determined, the features of intermediate layers should lie in a straight line between the data and the last-layer features because all features are free variables. Thus, minimizing the optimal transport regularizer can be reduced to minimize the distance between the data and the last-layer features, as shown in the following theorem. 
\begin{theorem}\label{thm3}
The problem in (\ref{prob}) is equivalent to
\begin{equation}\label{MUFM}
\min _{\boldsymbol{W}, \boldsymbol{H}} \mathcal{L}_{CE/MSE}(\boldsymbol{W},\boldsymbol{H},\boldsymbol{Y})+\frac{\lambda_{\boldsymbol{W}}}{2K}\|\boldsymbol{W}\|_{F}^{2}+\dfrac{\lambda}{2Kn}\|\boldsymbol{H}-\boldsymbol{X}\|_{F}^{2},
\end{equation}
where $\boldsymbol{X}$ is the input data matrix, $\boldsymbol{H}$ is the last-layer feature matrix for simplicity. 
\end{theorem}
\begin{IEEEproof}
$\forall k,i$, when $\boldsymbol{h}_{k, i}^0$ and $\boldsymbol{h}_{k, i}^L$ are fixed, we have
\begin{equation}
\begin{aligned}
\sum_{l=0}^{L-1}\left\|\boldsymbol{h}_{k, i}^{l+1}-\boldsymbol{h}_{k, i}^{l}\right\|_2^2&\geq \left\|\sum_{l=0}^{L-1}\left(\boldsymbol{h}_{k, i}^{l+1}-\boldsymbol{h}_{k, i}^{l}\right)\right\|_2^2\\
&=\left\|\boldsymbol{h}_{k, i}^{L}-\boldsymbol{h}_{k, i}^{0}\right\|_2^2
\end{aligned}
\end{equation}
the equality is achieved only when $\forall l$, 
\begin{equation}
\boldsymbol{h}_{k, i}^{l+1}-\boldsymbol{h}_{k, i}^{l}=\frac{1}{L}\left(\boldsymbol{h}_{k, i}^{L}-\boldsymbol{h}_{k, i}^{0}\right)
\end{equation}
which conclude the proof
\end{IEEEproof}

We call this model (\ref{MUFM}) the multilayer unconstrained feature model (MUFM). It is easy to show that, while taking data distribution into account, the global optima of MUFM is not consistent with NC. In fact, (\ref{MUFM}) can be rewritten as 
\begin{equation}\label{MUFM2}
\begin{aligned}
\min _{\boldsymbol{W}, \boldsymbol{H}} \quad&\mathcal{L}_{CE/MSE}(\boldsymbol{W},\boldsymbol{H},\boldsymbol{Y})+\frac{\lambda_{\boldsymbol{W}}}{2K}\|\boldsymbol{W}\|_{F}^{2} \\
&+\dfrac{\lambda}{2Kn}\|\boldsymbol{H}\|_{F}^{2} -\dfrac{\lambda}{Kn}\mathrm{Tr}(\boldsymbol{X}\boldsymbol{H}^{\top}). 
\end{aligned}
\end{equation}
By the results of UFM, the solution of (\ref{MUFM2}) without the last term $\frac{\lambda}{Kn}\mathrm{Tr}(\boldsymbol{X}\boldsymbol{H}^{\top})$ is the NC solution. But $\frac{\lambda}{Kn}\mathrm{Tr}(\boldsymbol{X}\boldsymbol{H}^{\top})$ makes $\boldsymbol{H}$ align with $\boldsymbol{X}$. To exhibit the collapse phenomenon in MUFM theoretically, we have

\begin{theorem}\label{coro1}
Suppose the initial data $\boldsymbol{X}$ is non-collapsed, then the optimal solutions of (\ref{MUFM}) are also non-collapsed. 
In addition, if the loss function is chosen to be MSE, the minimizer w.r.t. $\boldsymbol{W}$ has a closed-form expression $\boldsymbol{W}^*(\boldsymbol{H})=\boldsymbol{Y} \boldsymbol{H}^{\top}(\boldsymbol{H} \boldsymbol{H}^{\top}+n \lambda_{\boldsymbol{W}} \boldsymbol{I}_d)^{-1}$, which is a function of $\boldsymbol{H}$. If we substitute $\boldsymbol{W}^*(\boldsymbol{H})$ for $\boldsymbol{W}$, then there exists a large enough constant $C>0$, for any $\lambda>C$, the minimizer $\boldsymbol{H}_{1/\lambda}$ of (\ref{MUFM}) satisfies $\mathcal{PFC} _1(\boldsymbol{H}_{1/\lambda})<\mathcal{PFC} _1(\boldsymbol{X})$, which means the features are more collapsed than the data.
\end{theorem}

The proof of Theorem \ref{coro1} is analogous to the theorems in \cite{tirer2023perturbation}. For completeness, we provide it in Appendix \ref{appxC}. %In addition, empirical observation shows that the gap between $\|\boldsymbol{W}^*(\boldsymbol{H})\boldsymbol{H}-\boldsymbol{Y}\|_{F}^{2}$ and $\|\boldsymbol{WH}-\boldsymbol{Y}\|_{F}^{2}$ is relatively small in practical DNNs \cite{han2021neural}. Therefore, it shows that the optimal solutions of MUFM are not simplex ETF, but related to data. 
The coefficient of the optimal transport regularizer plays an essential role in the trade-off between the data and the ETF. If the coefficient is small, the solutions will be close to the solutions of UFM, i.e. the simplex ETF. If it is large, the solutions will be close to the data, but still more collapsed than the data. We give an empirical result to show the trade-off between the simplex ETF and data in Section \ref{sec4.2}. 
\begin{table}[!t]
\caption{The architectures of ResNet on various datasets}
\label{table1}
\centering
\renewcommand{\arraystretch}{1.1}
\setlength\tabcolsep{4pt}
\begin{tabular}{c|c|c|c|c|c}
\hline
Dataset & Data dim. & K & Intermediate dim. & \makecell[c]{\# of\\ blocks} & Layer \\
\hline
MNIST & $1\times28\times28$ & 1& 1000 & 10& FC\\
\hline
\multirow{2}{*}{\makecell[c]{Fashion\\MNIST}} & \multirow{2}{*}{$1\times28\times28$} & \multirow{2}{*}{2}& $32\times14\times14$ & 5 & \multirow{2}{*}{Conv}\\
\cline{4-5}
\multirow{2}{*}{}& \multirow{2}{*}{}&\multirow{2}{*}{}&$64\times7\times7$ & 4 & \multirow{2}{*}{} \\
\hline
\multirow{3}{*}{CIFAR10} & \multirow{3}{*}{$3\times32\times32$} & \multirow{3}{*}{3}& $64\times16\times16$ & 2 & \multirow{3}{*}{Conv}\\
\cline{4-5}
\multirow{3}{*}{}& \multirow{3}{*}{}&\multirow{3}{*}{}&$128\times8\times8$ & 3 & \multirow{3}{*}{} \\
\cline{4-5}
\multirow{3}{*}{}& \multirow{3}{*}{}&\multirow{3}{*}{}&$256\times4\times4$ & 4 & \multirow{3}{*}{} \\
\hline
\multirow{4}{*}{STL10} & \multirow{4}{*}{$3\times96\times96$} & \multirow{4}{*}{4}& $64\times48\times48$ & 5 & \multirow{4}{*}{Conv}\\
\cline{4-5}
\multirow{4}{*}{}& \multirow{4}{*}{}&\multirow{4}{*}{}&$128\times24\times24$ & 5 & \multirow{4}{*}{} \\
\cline{4-5}
\multirow{4}{*}{}& \multirow{4}{*}{}&\multirow{4}{*}{}&$256\times12\times12$ & 5 & \multirow{4}{*}{} \\
\cline{4-5}
\multirow{4}{*}{}& \multirow{4}{*}{}&\multirow{4}{*}{}&$512\times6\times6$ & 5 & \multirow{4}{*}{} \\
\hline
\multirow{4}{*}{CIFAR100} & \multirow{4}{*}{$3\times32\times32$} & \multirow{4}{*}{4}& $64\times32\times32$ & 3 & \multirow{4}{*}{Conv}\\
\cline{4-5}
\multirow{4}{*}{}& \multirow{4}{*}{}&\multirow{4}{*}{}&$128\times16\times16$ & 3 & \multirow{4}{*}{} \\
\cline{4-5}
\multirow{4}{*}{}& \multirow{4}{*}{}&\multirow{4}{*}{}&$256\times8\times8$ & 4 & \multirow{4}{*}{} \\
\cline{4-5}
\multirow{4}{*}{}& \multirow{4}{*}{}&\multirow{4}{*}{}&$512\times4\times4$ & 4 & \multirow{4}{*}{} \\
\hline
\end{tabular}
\end{table}

\section{Experiments}\label{sec4}
In this section, we first present the empirical evidence supporting the PFC conjecture when training ResNet on various datasets. We also show that these PFC metrics monotonically decrease along a straight line under the geodesic curve assumption. Then, we empirically show the distinctions between the optimal solutions of UFM and MUFM. The optimal solution of MUFM is not the simplex ETF and there exists a trade-off between data and simplex ETF by controlling the coefficient of the optimal transport regularizer $\lambda$.
\subsection{PFC Conjecture}\label{sec4.1}
To compare PFC metrics among different layer features, we need to keep the features in the same dimension (dim.). In this subsection, we first train a fully-connected (FC) ResNet whose intermediate dimension is set to $1000$ on the MNIST dataset. For the other datasets, such as Fashion MNIST, CIFAR10, STL10, and CIFAR100, we train the general convolutional (Conv) ResNets. These general ResNets gradually reduce the dimension to extract features, and we focus our investigation solely on the final $4$ or $5$ blocks whose input features are of the same dimension. The details of network architectures are presented in table \ref{table1}. The architectures are selected to satisfy that the empirical effective depth is equal to the actual depth since we only characterize PFC before effective depth. Thus, the last-layer features are NCC separable, indicating NC emerges.

Similar to the optimization setting in \cite{papyan2020prevalence}, we minimize the cross-entropy loss of each ResNet using SGD with $0.9$ momentum. The weight decay is set to $5\times 10^{-4}$ and the batch size is $128$. We train the network for 350 epochs. The initial learning rate is set to $0.005$ for MNIST and $0.01$ for other datasets, and it is divided by $10$ at $117$-th and $233$-rd epochs.

To demonstrate PFC conjecture empirically, we consider the following metrics in each layer to measure PFC,

(PFC1) Variability collapse:
\begin{equation}
\mathcal{PFC} _1(\boldsymbol{H}^{l}):= \frac{\mathbb{E}\left[\left\|\boldsymbol{h}_{k,i}^l-\boldsymbol{h}_{k}^l\right\|_2^2\right]}{\mathbb{E}\left[\left\|\boldsymbol{h}_{k}^l-\boldsymbol{h}_G^l\right\|_2^2\right]} .
\end{equation}

(PFC2) Convergence to simplex ETF:
\begin{equation}
\mathcal{PFC} _2(\boldsymbol{H}^{l}):=\left\|\frac{\tilde{\boldsymbol{H}^{l}}^{\top} \tilde{\boldsymbol{H}^{l}}}{\left\|\tilde{\boldsymbol{H}^{l}}^{\top} \tilde{\boldsymbol{H}^{l}}\right\|_F}-\boldsymbol{E}\right\|_F,
\end{equation}
where $
\tilde{\boldsymbol{H}^{l}}:=\left[
\boldsymbol{h}_1^{l}-\boldsymbol{h}_G^{l}, \cdots, \boldsymbol{h}_K^{l}-\boldsymbol{h}_G^{l}
\right] \in \mathbb{R}^{d \times K}
$ is the centered class mean matrix, $\boldsymbol{E}=\frac{1}{\sqrt{K-1}}\left(\boldsymbol{I}_K-\frac{1}{K} \boldsymbol{1}_K \boldsymbol{1}_K^{\top}\right)$.

(PFC3) NCC accuracy:
\begin{equation}
\mathcal{PFC} _3(\boldsymbol{H}^{l}):=\frac{1}{nK}\sum_{k=1}^{K}\sum_{i=1}^{n} \mathbb{I}\left[\underset{c \in[K]}{\arg \min }\left\|\boldsymbol{h}_{k,i}^l-\boldsymbol{h}_{c}^l\right\|_2 = k\right].
\end{equation}

\begin{figure*}[!t]
\centering
\includegraphics[width=0.94\textwidth]{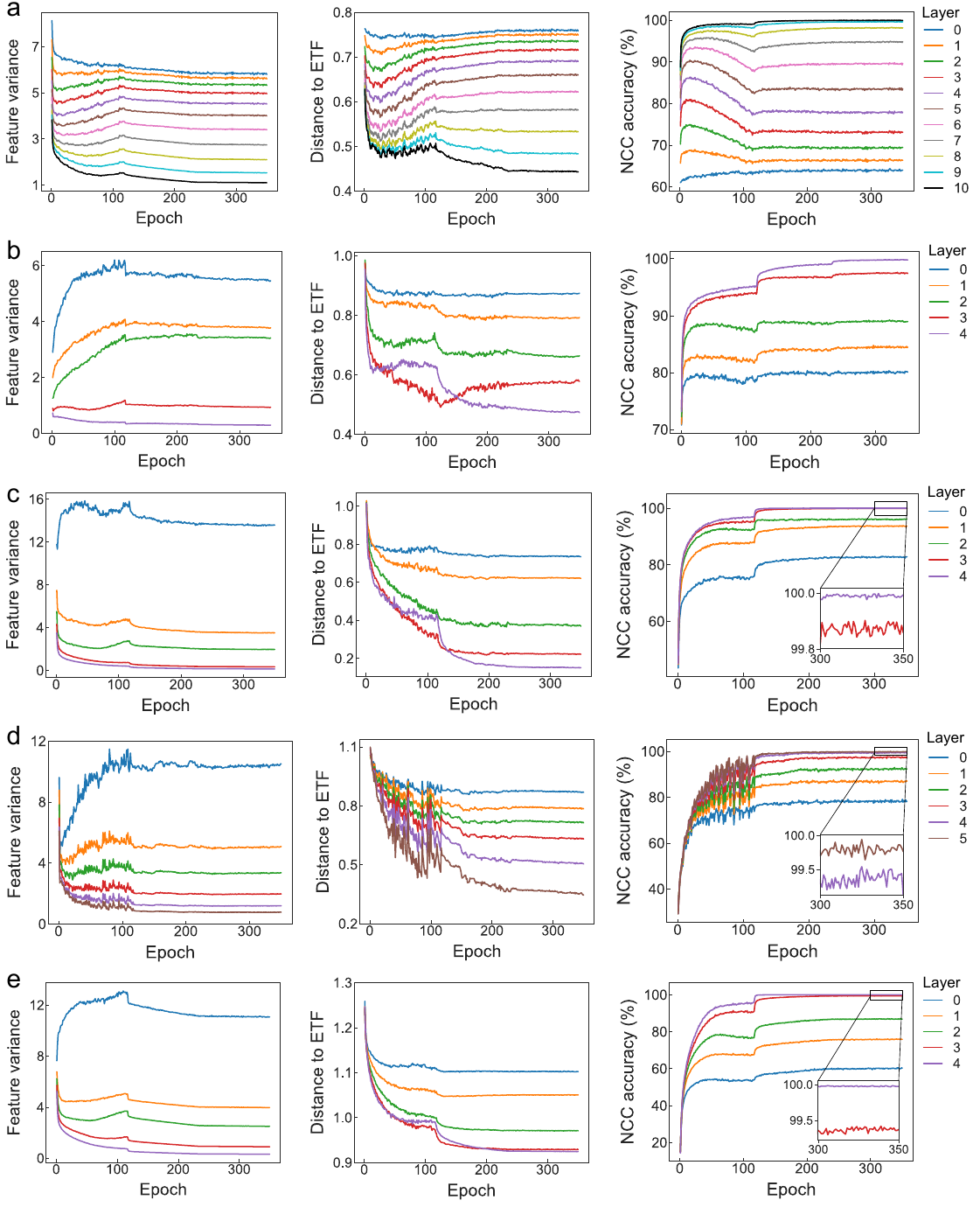}
\caption{\textbf{Features of ResNet exhibiting PFC on various datasets.} In each row, the PFC metrics of different layers on MNIST, Fashion MNIST, CIFAR10, STL10, and CIFAR100 are plotted (a-e). The `layer0' represents the input features.} 
\label{PFC1} 
\end{figure*}

\begin{figure*}[!t]
\centering 
\includegraphics[width=0.88\textwidth]{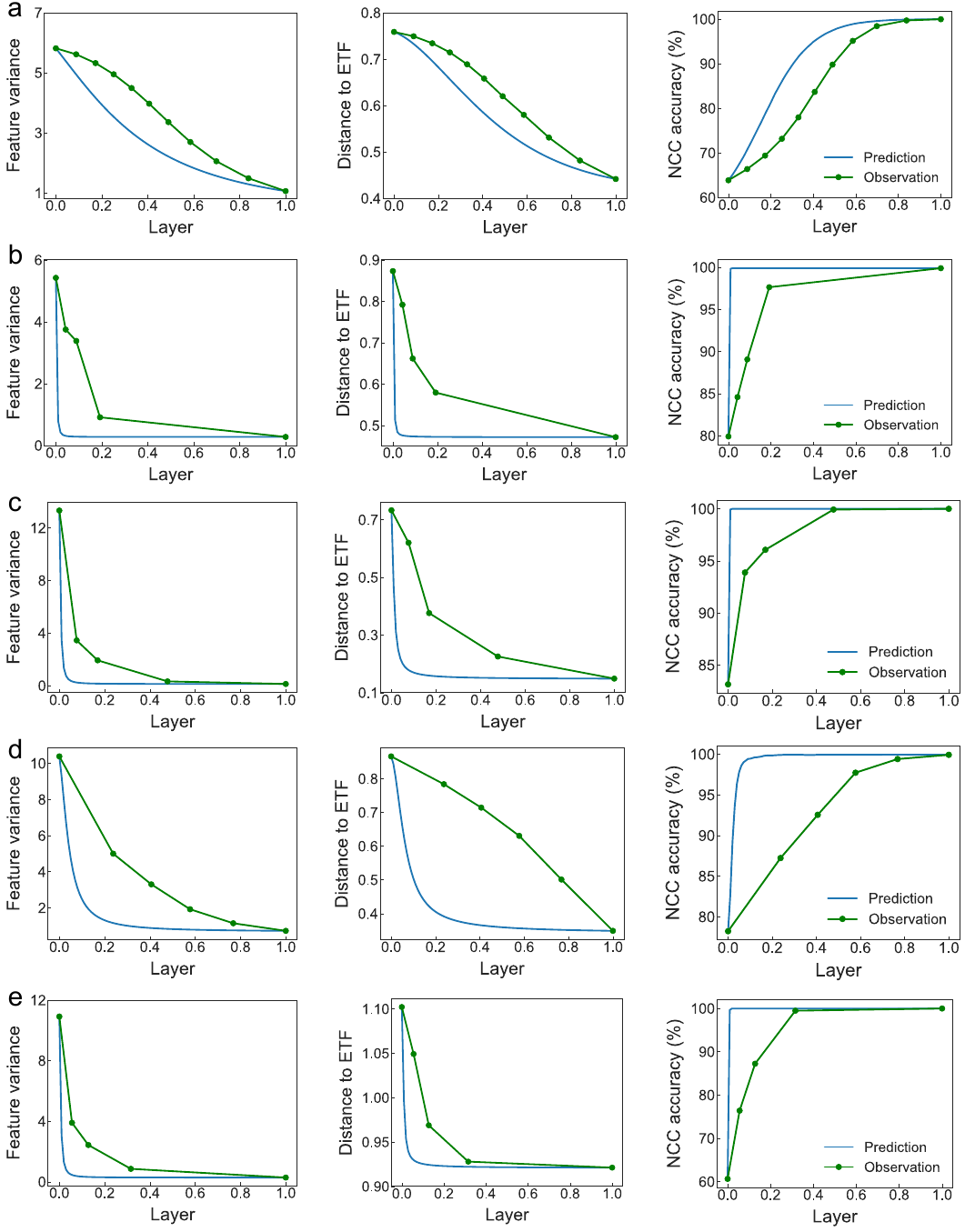}
\caption{\textbf{PFC on various datasets at the final epoch.} The green points are calculated by the features of different layers and connected by a green line. The blue curve represents the predicted values under the geodesic curve assumption. The abscissa represents the relative position where $0$ corresponds to the input features. Both observations and predictions are monotonic on various datasets (a)-(e) correspond to the same datasets as in Figure \ref{PFC1}.}
\label{PFC2}
\end{figure*}

We illustrate the evolution of the PFC metrics during training on various datasets (Figure \ref{PFC1}). To show the degree of collapse at each layer more clearly, we plot the PFC metrics of each layer at the final epoch (Figure \ref{PFC2}). The horizontal axis represents their relative position, scaled to the range of $[0,1]$. Formally, for $l$-layer features $\{\boldsymbol{h}_{k, i}^{l}\}$, their coordinate is calculated as follows: 
$$\dfrac{\sum\limits_{l=0}^{l-1} \sum\limits_{k=1}^K \sum\limits_{i=1}^n \left\|\boldsymbol{h}_{k, i}^{l+1}-\boldsymbol{h}_{k, i}^l\right\|_2}{\sum\limits_{l=0}^{L-1} \sum\limits_{k=1}^K \sum\limits_{i=1}^n \left\|\boldsymbol{h}_{k, i}^{l+1}-\boldsymbol{h}_{k, i}^l\right\|_2}.
$$
Additionally, the geodesic curve assumption allows us to predict the metrics along the line $\boldsymbol{h}_{k, i}^{(t)}=(1-t) \boldsymbol{h}_{k, i}^{(0)}+t \boldsymbol{h}_{k, i}^{(1)}, t \in [0,1]$, which connects the input features $\{\boldsymbol{h}_{k, i}^{(0)}\}$ and the last-layer features $\{\boldsymbol{h}_{k, i}^{(1)}\}$. For each $\{\boldsymbol{h}_{k, i}^{(t)}\}, t \in [0,1]$, we compute the PFC metrics and connect them (blue curve in Figure \ref{PFC2}). Empirical evidence supports the PFC conjecture when training ResNets. More concretely, before effective depth, features of each layer tend to collapse to their class means progressively, (centered) class means of each layer tend to collapse to the simplex ETF progressively, and NCC accuracy rates of each layer tend to collapse to 1 progressively. Under the geodesic curve assumption, the metrics monotonically decrease along the straight line. For each layer, these metrics start at random initialization and converge to PFC at the final epoch.

\subsection{Numerical Experiments in MUFM}\label{sec4.2}
\begin{figure*}[htbp]
\centering
\includegraphics[width=0.9\textwidth]{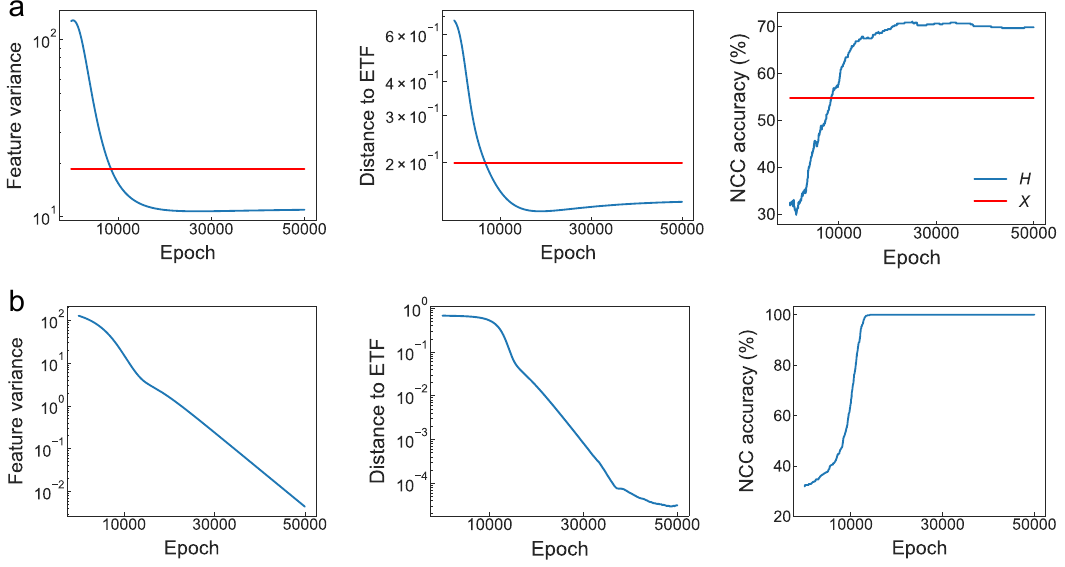}
\caption{\textbf{The difference between MUFM and UFM in terms of global optimality.} \textbf{(a):} The collapse degree of the solution of MUFM, measured by PFC metrics. The red line is the collapse degree of data.
\textbf{(b):} The collapse degree of the solution of UFM, measured by PFC metrics (log-scale). The solution of UFM exhibits NC but MUFM doesn't.}
\label{MUMF1}
\end{figure*}

\begin{figure*}[htbp]
\centering
\includegraphics[width=0.9\textwidth]{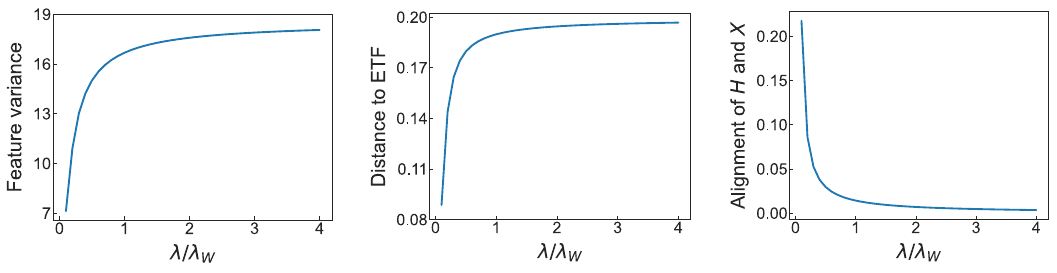}

\caption{\textbf{Trade-off in MUFM.} The abscissa is the ratio $\dfrac{\lambda}{\lambda_{\boldsymbol{W}}}$, we keep the value of $\lambda_{\boldsymbol{W}}$ and only vary $\lambda$. The first two figures show that increasing $\lambda$ makes the solution of MUFM away from the NC solution. The last figure shows that increasing $\lambda$ makes the solution of MUFM close to the data.}
\label{MUMF2}
\end{figure*}

%In section \ref{aba:sec3.3}, we propose a new model MUFM which is more general than UFM, and show that the global optima of MUFM is different from that of UFM. 
In this subsection, we numerically solve both MUFM and UFM. The solutions of both UFM and MUFM exhibit the NC phenomenon, while the NC in MUFM is slighter (Figure \ref{MUMF1}). Furthermore, MUFM shows the trade-off between the simplex ETF and data while changing the coefficient of optimal transport regularizer $\lambda$ (Figure \ref{MUMF2}).

Consider MUFM (\ref{MUFM}) in the setting $k=5, d=20, n=100, \lambda_{\boldsymbol{W}}=0.005, \lambda=0.001$, $\boldsymbol{W}$ and $\boldsymbol{H}$ are initialized with the standard normal distribution. The loss function is chosen to be MSE. The data $\{\boldsymbol{x}_{k,i}\}_{i=1}^n$ for each class $k$ are sampled from a Gaussian random vector $\mathcal{N}(\boldsymbol{\mu}_k,\boldsymbol{I}_d)$, where $\boldsymbol{\mu}_k\in\mathbb{R}^d$ is uniformly sampled from the unit sphere. UFM is in the same setting without sampling $\boldsymbol{X}$. We solve the problems by plain gradient descent with a learning rate $0.1$ for $5\times 10^4$ epochs. We then measure the solution $\boldsymbol{H}$ specifically by PFC metrics to show if it is consistent with NC. (Original NC includes self-duality: the alignment of the classifier and the last-layer features, we will not measure this additionally). We can see that the solution of UFM is much closer to the NC solution (each metric is in a smaller magnitude), but the solution of MUFM is not (Figure \ref{MUMF1}). Theorem \ref{coro1} tells us that $\mathcal{PFC} _1(\boldsymbol{H})<\mathcal{PFC} _1(\boldsymbol{X})$ for large enough $\lambda$, we compute all PFC metrics of the data $\boldsymbol{X}$ and show that the collapse degree (not only $\mathcal{PFC}_1$, but all PFC metrics) of solution $\boldsymbol{H}$ is larger than data empirically.

To show the trade-off between data and simplex ETF, we gradually vary the optimal transport regularizer coefficient $\lambda$ from $0.0005$ to $0.02$, while keeping other parameters constant. For each value of $\lambda$, we solve the MUFM as above and measure the collapse degree of $\boldsymbol{H}$ by PFC metrics. In addition, we measure the alignment of the solution $\boldsymbol{H}$ and the data $\boldsymbol{X}$ by 
$$
\left \| \dfrac{\boldsymbol{H}}{\| \boldsymbol{H}\|_F} - 
\dfrac{\boldsymbol{X}}{ \|\boldsymbol{X}\|_F} \right\|_F.
$$
We illustrate that for small values of the coefficient $\lambda$, the solution tends to be close to the simplex ETF (Figure \ref{MUMF2}). Conversely, for large values of $\lambda$, the solution moves away from the ETF and becomes closer to the initial data.


\section{Related work}\label{sec5}
\textbf{NC and (extended) UFM.} The NC phenomenon has been described in \cite{papyan2020prevalence,han2021neural} and proved to emerging by UFM under various settings. Mixon \textit{et al.} \cite{mixon2020neural}, Han \textit{et al.} \cite{han2021neural} proved that the dynamics of SGD converge to the NC solutions. Several studies proved that the global minimizers are NC solutions \cite{lu2020neural,weinan2022emergence,fang2021exploring,ji2021unconstrained,tirer2022extended,zhu2021geometric,zhou2022optimization,yaras2022neural,zhou2022all}. Moreover, some studies \cite{zhu2021geometric,zhou2022optimization,yaras2022neural,zhou2022all} proved that UFM has benign loss landscape. Beyond UFM, Tirer and Bruna \cite{tirer2022extended} extended UFM to two layers, and characterized its global optimality. Dang \textit{et al.} \cite{dang2023neural} extended UFM to multiple linear layers and characterized the global optimality. S{\'u}ken{\'\i}k \textit{et al.} \cite{sukenik2024deep} extended UFM to arbitrary non-linear layers and proved that after a certain layer, the optimal solutions for binary classification exactly exhibit NC. Since UFM doesn't consider the effect of data, in our paper, we propose a more general model MUFM which models the cases when NC is inaccurate. Tirer \textit{et al.} \cite{tirer2023perturbation} also proposed a new model to analyze the ``near collapse" situation which is similar to MUFM. But MUFM is constructed more reasonably: replacing the weight decay with the optimal transport regularizer. We claim that the regularizer on $\boldsymbol{H}$ is unnecessary to model DNNs, which is not used in practice. Yang \textit{et al.} \cite{pmlr-v202-yang23m} also consider the role of the initial data in NC, showing that the collapsed features corresponding to each label can exhibit fine-grained structures determined by the initial data. This provides experimental evidence to MUFM in modeling the alignment between the input data and features.

\textbf{NC at intermediate layers.} The NC phenomenon only characterizes the last-layer features. Several studies tried to extend it to intermediate layers. Many empirical results show that NC1 emerges at intermediate layers but progresses weaker than the last layer \cite{hui2022limitations,galanti2021role,ben2022nearest,MinimalDepth,tirer2023perturbation,he2023law}. Ben-Shaul and Dekel \cite{ben2022nearest}, Galanti \textit{et al.} \cite{MinimalDepth} also provided a similar result of NC4. These empirical works did not explore all NC properties, Rangamani \textit{et al.} \cite{rangamani2023feature} extended all NC properties to intermediate layers including NC3 (feature-weight alignment). However, Rangamani \textit{et al.} \cite{rangamani2023feature} did not characterize the evolution of the NC properties; instead, it showed that there exists a hidden layer beyond which all subsequent layers exhibit the NC properties. They provided the NC properties after effective depth. Parker \textit{et al.} \cite{parker2023neural} provided similar empirical results in different settings to characterize the intermediate layers. Our PFC conjecture focuses on the layers of ResNet before effective depth. Moreover, we show the monotonicity of collapse degree and give some theoretical understandings.

Tirer \textit{et al.} \cite{tirer2023perturbation} also provided a theoretical result that the PFC1 metric monotonically decreases along a gradient flow. Thus, the PFC1 metric will decrease across depth when the network is optimized layer-wisely (a new layer is added on the top of the trained network each time, i.e., cascade learning). However, this theoretical result does not match the more common empirical observation in \cite{hui2022limitations,galanti2021role,ben2022nearest,MinimalDepth} perfectly, whose training framework is end-to-end. Our result shows this monotonic decrease of PFC1 metric across depth directly for end-to-end training with some mild assumptions.

\section{Conclusion and Discussion}\label{sec6}
NC is an intriguing phenomenon observed during the training of DNNs, revealing insights into the geometry of the last-layer features and classifier, thereby elucidating the efficacy of DNNs for classification tasks. In this work, we extend the NC phenomenon to the intermediate layers of ResNet and propose a novel conjecture termed PFC indicated by PFC1, PFC2 and PFC3. We provide empirical evidence to support the PFC conjecture. In theory, we introduce the geodesic curve assumption and prove that the metrics of PFC indeed decrease monotonically across depth at the terminal phase of training under some assumptions. The PFC conjecture suggests that the ResNet can progressively concentrate the features and separate the class means to achieve classification problems.

The geodesic curve assumption also sheds light on extending UFM to MUFM. Taking the intermediate layer into account, we propose a new surrogate model MUFM to understand NC and PFC. This model establishes a connection between the data and the last-layer features through the optimal transport regularizer. Then, the optimal solution of MUFM is inconsistent with NC, but the features are more concentrated than the input data. The optimal transport regularizer coefficient $\lambda$ can control the distance between the solution and the data. Compared to UFM, MUFM models the terminal phase of DNN more comprehensively.

Further research could explore more theoretical aspects of MUFM. Current PFC conjecture and theoretical understanding focus on ResNet. It will be interesting to characterize the forward propagation of other DNN architectures. Additionally, we hope that the PFC conjecture could inspire the design of new loss functions or architectures aimed at improving classification performance.

\section*{Acknowledgments}
The authors would like to thank Zhenpeng Man, Tianyu Ruan, and Xiayang Li for their helpful discussion.

\bibliographystyle{IEEEtran}
\bibliography{IEEEabrv}

{\appendices
\section{Proof of Theorem \ref{thm1}}\label{appxA}
\begin{IEEEproof}
The PFC1 metric equals zero at $t = 1$ because $\{\boldsymbol{h}_{k, i}^{(1)}\}$ exhibits NC.
%We first consider PFC1,
Recall the metric of PFC1 is
\begin{equation*}
\mathcal{PFC}_1(t) = \frac{\mathbb{E}\left[\left\|\boldsymbol{h}_{k,i}^{(t)}-\boldsymbol{h}_{k}^{(t)}\right\|_2^2\right]}{\mathbb{E}\left[\left\|\boldsymbol{h}_{k}^{(t)}-\boldsymbol{h}_G^{(t)}\right\|_2^2\right]} .
\end{equation*} 

By (\ref{line}), the numerator can be rewritten as
\begin{equation*}
\begin{aligned}
\mathbb{E}\left[\left\|\boldsymbol{h}_{k,i}^{(t)}-\boldsymbol{h}_{k}^{(t)}\right\|_2^2\right] &= \mathbb{E}\left[\left\|(1-t)\left(\boldsymbol{h}_{k,i}^{(0)}-\boldsymbol{h}_{k}^{(0)}\right)\right\|_2^2\right]
\\
&=(1-t)^2 \mathbb{E}\left[\left\|\boldsymbol{h}_{k,i}^{(0)}-\boldsymbol{h}_{k}^{(0)}\right\|_2^2\right].
\end{aligned}
\end{equation*}

For the denominator, denote $\tilde{\boldsymbol{h}}_{k}^{(t)} = \boldsymbol{h}_{k}^{(t)}-\boldsymbol{h}_G^{(t)}$, then 
\begin{equation*}
\begin{aligned}
\dfrac{d\mathbb{E}\left[\left\|\boldsymbol{h}_{k}^{(t)}-\boldsymbol{h}_G^{(t)}\right\|_2^2\right]}{dt} 
&=\dfrac{d}{dt} \left( \frac{1}{K}\sum_{k=1}^{K}\left\|\tilde{\boldsymbol{h}}_{k}^{(t)}\right\|_2^2\right)
\\
&= \frac{1}{K}\sum_{k=1}^{K}\dfrac{d}{dt} \left\|\tilde{\boldsymbol{h}}_{k}^{(t)}\right\|_2^2
\\
&= \frac{2}{K}\sum_{k=1}^{K}\langle\tilde{\boldsymbol{h}}_{k}^{(t)},\dfrac{d\tilde{\boldsymbol{h}}_{k}^{(t)}}{dt}\rangle
\\
&= \frac{2}{K}\sum_{k=1}^{K}\langle\tilde{\boldsymbol{h}}_{k}^{(t)},\tilde{\boldsymbol{h}}_{k}^{(1)}-\tilde{\boldsymbol{h}}_{k}^{(0)}\rangle.
\end{aligned}
\end{equation*}

Now, compute the derivative of $\mathcal{PFC}_1(t)$ w.r.t. $t$, we have
\begin{equation*}
\begin{aligned}
\dfrac{d\mathcal{PFC}_1(t)}{dt} 
&=
\dfrac{(-2+2t)\mathbb{E}\left[\left\|\boldsymbol{h}_{k,i}^{(0)}-\boldsymbol{h}_{k}^{(0)}\right\|_2^2\right]\mathbb{E}\left[\left\|\tilde{\boldsymbol{h}}_{k}^{(t)}\right\|_2^2\right] }
{\left(\mathbb{E}\left[\left\|\tilde{\boldsymbol{h}}_{k}^{(t)}\right\|_2^2\right]\right)^2} \\
&\quad-\dfrac{\mathbb{E}\left[\left\|\boldsymbol{h}_{k,i}^{(t)}-\boldsymbol{h}_{k}^{(t)}\right\|_2^2\right]\dfrac{d\mathbb{E}\left[\left\|\boldsymbol{h}_{k}^{(t)}-\boldsymbol{h}_G^{(t)}\right\|_2^2\right]}{dt} }{\left(\mathbb{E}\left[\left\|\tilde{\boldsymbol{h}}_{k}^{(t)}\right\|_2^2\right]\right)^2}
\\
&= 
- \dfrac{2(1-t)\mathbb{E}\left[\left\|\boldsymbol{h}_{k,i}^{(0)}-\boldsymbol{h}_{k}^{(0)}\right\|_2^2\right]}{K\left(\mathbb{E}\left[\left\|\tilde{\boldsymbol{h}}_{k}^{(t)}\right\|_2^2\right]\right)^2}\cdot \left(\sum_{k=1}^{K}\left\|\tilde{\boldsymbol{h}}_{k}^{(t)}\right\|_2^2\right. \\
&\quad\left. + (1-t) \sum_{k=1}^{K}\langle\tilde{\boldsymbol{h}}_{k}^{(t)},\tilde{\boldsymbol{h}}_{k}^{(1)}-\tilde{\boldsymbol{h}}_{k}^{(0)}\rangle\right)
\\
&= -
\dfrac{2(1-t)\mathbb{E}\left[\left\|\boldsymbol{h}_{k,i}^{(0)}-\boldsymbol{h}_{k}^{(0)}\right\|_2^2\right]}{K\left(\mathbb{E}\left[\left\|\tilde{\boldsymbol{h}}_{k}^{(t)}\right\|_2^2\right]\right)^2}\cdot
\left(t\sum_{k=1}^{K}\left\|\tilde{\boldsymbol{h}}_{k}^{(1)}\right\|_2^2 \right.\\
&\left. \quad+ (1-t)\sum_{k=1}^{K}\langle\tilde{\boldsymbol{h}}_{k}^{(0)},\tilde{\boldsymbol{h}}_{k}^{(1)}\rangle
\right).
\end{aligned}
\end{equation*}
Based on the assumption 
\begin{equation*}
\sum_{k=1}^{K}\langle \boldsymbol{h}_{k}^{(0)} - \boldsymbol{h}_{G}^{(0)},\boldsymbol{h}_{k}^{(1)} - \boldsymbol{h}_{G}^{(1)}\rangle = \sum_{k=1}^{K}\langle\tilde{\boldsymbol{h}}_{k}^{(0)},\tilde{\boldsymbol{h}}_{k}^{(1)}\rangle \ge 0,
\end{equation*}
$\dfrac{d\mathcal{PFC}_1(t)}{dt} < 0$ for $\forall t\in[0,1)$ and thus $\mathcal{PFC}_1$ strictly decreases to zero in $t\in[0,1]$.

\end{IEEEproof}

\section{Proof of Theorem \ref{thm2}}\label{appxB}
\begin{IEEEproof}
The PFC2 metric equals zero at $t = 1$ because the end point is the global minimum, exhibiting NC. For any start point $\tilde{\boldsymbol{H}}(0)$, we turn to prove the PFC2 metric monotonically decreases at a neighborhood
$(1-\delta, 1]$.

Let’s consider a surrogate metric that has the same monotonicity as the PFC2 metric, 
\begin{equation*}
\begin{aligned}
\tilde{\mathcal{PFC}_2}(t):&= \frac{1}{2}\left\|\frac{\tilde{\boldsymbol{H}}(t)^{\top} \tilde{\boldsymbol{H}}(t)}{\left\|\tilde{\boldsymbol{H}}(t)^{\top} \tilde{\boldsymbol{H}}(t)\right\|_F}-\boldsymbol{E}\right\|_F^2\\
& = \dfrac{\frac{1}{2}\left\|\tilde{\boldsymbol{H}}(t)^{\top} \tilde{\boldsymbol{H}}(t)-\left\|\tilde{\boldsymbol{H}}(t)^{\top} \tilde{\boldsymbol{H}}(t)\right\|_F\boldsymbol{E}\right\|_F^2}
{\left\|\tilde{\boldsymbol{H}}(t)^{\top} \tilde{\boldsymbol{H}}(t)\right\|_F^2}\\
& \triangleq \dfrac{f(t)}{g(t)},
\end{aligned}
\end{equation*}
where $\boldsymbol{E}=\frac{1}{\sqrt{K-1}}\left(\boldsymbol{I}_K-\frac{1}{K} \boldsymbol{1}_K \boldsymbol{1}_K^{\top}\right)$, $f$, $g$ are polynomial functions by (\ref{line2}).

It is easy to get that $f(1)=0$, $g(t)>0$ by the chain rule, 
\begin{equation*}
f'(t)= \left\langle\tilde{\boldsymbol{H}}(t)^{\top} \tilde{\boldsymbol{H}}(t)-\left\|\tilde{\boldsymbol{H}}(t)^{\top} \tilde{\boldsymbol{H}}(t)\right\|_F\boldsymbol{E}, \boldsymbol{D}(t)
\right\rangle, 
\end{equation*}
where 
\begin{equation*}
\boldsymbol{D}(t) =\frac{d\left(\tilde{\boldsymbol{H}}(t)^{\top} \tilde{\boldsymbol{H}}(t)-\left\|\tilde{\boldsymbol{H}}(t)^{\top} \tilde{\boldsymbol{H}}(t)\right\|_F\boldsymbol{E}\right)}{dt}
\end{equation*}
thus, $f'(1) = 0$ since $\tilde{\boldsymbol{H}}(1)^{\top} \tilde{\boldsymbol{H}}(1)=\left\|\tilde{\boldsymbol{H}}(1)^{\top} \tilde{\boldsymbol{H}}(1)\right\|_F\boldsymbol{E}$, and we have 
\begin{equation*}
f''(1) = \left\|\boldsymbol{D}(1)\right\|_F^2>0.
\end{equation*}
Then, the first derivative of $\tilde{\mathcal{PFC}_2}(t)$ at $t=1$ is
\begin{equation*}
\left.\dfrac{d\tilde{\mathcal{PFC}_2}(t)}{dt} \right|_{t=1}= \left.\dfrac{f'(t)g(t)-f(t)g'(t)}{g^2(t)}\right|_{t=1} =0.
\end{equation*}
The second derivative of $\tilde{\mathcal{PFC}_2}(t)$ at $t=1$ is
\begin{equation*}
\left.\dfrac{d^2\tilde{\mathcal{PFC}_2}(t)}{dt^2} \right|_{t=1}= \dfrac{f''(1)}{g(1)} > 0.
\end{equation*}

Since the second derivative is continuous, there exists a neighborhood $(1-\delta, 1]$, s.t. $\dfrac{d^2\tilde{\mathcal{PFC}_2}(t)}{dt^2}>0$. Thus, the first derivative is a monotonic increase in the neighborhood, i.e., $\dfrac{d\tilde{\mathcal{PFC}_2}(t)}{dt}<0$ in $(1-\delta, 1]$. 

Suppose $\tilde{\boldsymbol{H}}(1-\delta)$ is the start point, the theorem is proved.
\end{IEEEproof}

\section{Proof of Theorem \ref{coro1}}\label{appxC}
\begin{IEEEproof}
The first statement is trivial. For the second, recall that the minimizer w.r.t. $\boldsymbol{W}$ in (\ref{MUFM}) has a closed-form expression $\boldsymbol{W}^*(\boldsymbol{H})=\boldsymbol{Y} \boldsymbol{H}^{\top}(\boldsymbol{H} \boldsymbol{H}^{\top}+n \lambda_{\boldsymbol{W}} \boldsymbol{I}_d)^{-1}$. Substitute $\boldsymbol{W}^*(\boldsymbol{H})$ for $\boldsymbol{W}$ in (\ref{MUFM}), the optimization problem for $\boldsymbol{H}$ is 
\begin{equation}\label{central path}
\min _{\boldsymbol{H}} \mathcal{L}(\boldsymbol{H}) +\dfrac{\lambda}{2Kn}\|\boldsymbol{H}-\boldsymbol{X}\|_{F}^{2},
\end{equation}
where $\mathcal{L}(\boldsymbol{H}) := \frac{1}{2K n} \left\|\boldsymbol{W}^*(\boldsymbol{H})\boldsymbol{H}-\boldsymbol{Y}\right\|_{F}^{2}+\frac{\lambda_{\boldsymbol{W}}}{2K}\|\boldsymbol{W}^*(\boldsymbol{H})\|_{F}^{2}$.
Denote the minimizer of (\ref{central path}) for certain $\lambda$ is $\boldsymbol{H}_{1 / \lambda}$, which satisfies the first order optimality condition of (\ref{central path}):
\begin{equation*}
K n \nabla \mathcal{L}\left(\boldsymbol{H}_{1 / \lambda}\right) + \frac{\boldsymbol{H}_{1 / \lambda}-\boldsymbol{X}}{1 / \lambda}= 0.
\end{equation*}
Let $\boldsymbol{H}_0 = \boldsymbol{X}$, and obviously, for $\lambda \rightarrow \infty$, then $\boldsymbol{H}_{1 / \lambda} \rightarrow \boldsymbol{H}_0$. Thus, rewriting $1 / \lambda$ as $t$, 
\begin{equation*}
\left.\frac{d \boldsymbol{H}_t}{d t}\right|_{t=0}=\lim _{\lambda \rightarrow \infty} \frac{\boldsymbol{H}_{1 / \lambda}-\boldsymbol{H}_0}{1 / \lambda}=-K n \nabla \mathcal{L}\left(\boldsymbol{H}_0\right).
\end{equation*}
Then for large enough $\lambda$, the minimizer of (\ref{central path}) can be approximated by the gradient flow
\begin{equation}\label{GD}
\frac{d \boldsymbol{H}_t}{d t}=-K n \nabla \mathcal{L}\left(\boldsymbol{H}_t\right)
\end{equation}

The following lemma shows that $\mathcal{PFC}_1(\boldsymbol{H}_t)$ monotonically decreases along this flow, which is a corollary of Theorem 4.1 in \cite{tirer2023perturbation}.
\begin{lemma}\label{lemma1}
Assume that $\boldsymbol{H}_0$ is non-collapsed, along the gradient flow (\ref{GD}), we have that, $t \mapsto \mathrm{Tr}\left(\boldsymbol{\Sigma}_W\left(\boldsymbol{H}_t\right)\right)$ decreases and $t \mapsto \mathrm{Tr}\left(\boldsymbol{\Sigma}_B\left(\boldsymbol{H}_t\right)\right)$ strictly increases. Thus, $\mathcal{PFC}_1(\boldsymbol{H}_t)$ strictly decreases w.r.t $t$ until $t$ reaches zero.
\end{lemma}

As a consequence of lemma \ref{lemma1}, there exists a large enough constant $C>0$, then for any $\lambda>C$, $\mathcal{PFC}_1(\boldsymbol{H}_{1/\lambda})<\mathcal{PFC}_1(\boldsymbol{H}_0)=\mathcal{PFC}_1(\boldsymbol{X}).$
\end{IEEEproof}
}

\end{document}